# Threat analysis of IoT networks Using Artificial Neural Network Intrusion Detection System

Elike Hodo, Xavier Bellekens, Andrew Hamilton, Pierre-Louis Dubouilh,
Ephraim Iorkyase, Christos Tachtatzis and Robert Atkinson
Department of Electronic & Electrical Engineering
University of Strathclyde
Glasgow, G1 1XW, UK
E-mail :{ name.surname}@strath.ac.uk

*Abstract*— The Internet of things (IoT) is still in its infancy and has attracted much interest in many industrial sectors including medical fields, logistics tracking, smart cities and automobiles. However as a paradigm, it is susceptible to a range of significant intrusion threats. This paper presents a threat analysis of the IoT and uses an Artificial Neural Network (ANN) to combat these threats. A multi-level perceptron, a type of supervised ANN, is trained using internet packet traces, then is assessed on its ability to thwart Distributed Denial of Service (DDoS/DoS) attacks. This paper focuses on the classification of normal and threat patterns on an IoT Network. The ANN procedure is validated against a simulated IoT network. The experimental results demonstrate 99.4% accuracy and can successfully detect various DDoS/DoS attacks.

I. INTRODUCTION

The internet of things (IoT) is a network of distributed (sensor) nodes, (cloud) servers, and software. This paradigm permits measurands to be sensed and processed at in real-time creating a direct interaction platform between cyber-physical systems. Such an approach leads to improved efficiency in the generation and usage of data leading to economic benefits [1].

Research conducted by Cisco reports there are currently 10 billion devices connected, compared to the world population of over 7 billion and it is believed it will increase by 4% by the year 2020 [2].

Threats to the IoT paradigm are on the rise; however patterns within recorded data can be analyzed to help predict threats [3]. Intruder types are categorized into two:
1) External Intruders – these are people who fall outside the network and hence do not have permissions on the network. They operate by sending malware, or by using exploits to gain access to systems [4].
2) Internal Intruders – these people have rights and privileges to access the network, but misuse them malevolently. These types of attack include changing important data content or theft of confidential data. All these threats can be done physically by hacking into the computer system or by accessing a network remotely without permission [4].

IoT threat can be classified into four types [5]:
1) Denial of Service (DoS) – This threat denies or prevents user's resource on a network by introducing useless or unwanted traffic
2) Malware – Attackers use executable code to disrupt devices on the IoT network. They may gather sensitive information, or gain unauthorized access to the devices. The attacker can take advantage of flaws in the firmware running on the devices and run their software to disrupt the IoT architecture.
3) Data breaches – This is a security incident where sensitive, protected or confidential data is retrieved from the network. Attackers can spoof ARP packets to listen on the communication between peers on the network.
4) Weakening Perimeters – IoT network devices are currently not designed considering the pervasive security. Network security mechanisms are not often present in the devices making the network a vulnerable one for threats [5][6].

An intrusion detection system (IDS) is a security detection system put in place to monitor networks and computer systems. It has been in existence since the 1980s [7]. Previous and recent works using Artificial Neural network intrusion detection system on KDD99 data set [8], [9],[10],[11] show a promising performance for intrusion detection.

In this paper an ANN is used as an offline IDS to gather and analyze information from various part of the IoT network and identify a DoS attack on the network.

The rest of the paper is organized as follows: Section II gives a review of intrusion detection system. Section III introduces the learning procedure of Artificial Neural Network algorithm. Section IV gives a description of network architecture. Finally section V presents results analysis, future works and conclusion are presented in section VI.

## II. INTRUSION DETECTION

An IDS is strategically placed on a network to detect threats and monitor packets. The functions of the IDS include offering information on threats, taking corrective steps when it detects threats and taking record of all events within a network [12]

### A. Intrusion Detection Classification

Table I. Comparison of HIDS and NIDS performance [13]

| Performance in terms of: | Host-Based IDS | Network-Based IDS |
|---|---|---|
| Intruder deterrence | Strong deterrence for inside intruders | Strong deterrence for outside intruders |
| Threat response time | Weak real time response but performs better for a long term attack | Strong response time against outside intruders |
| Assessing damage | Excellent in determining extent of damage | Very weak in determining extent of damage |
| Intruder prevention | Good at preventing inside intruders | Good at preventing outside intruders |
| Threat anticipation | Good at trending and detecting suspicious behavior patterns | Good at trending and detecting suspicious behavior patterns |

Intrusion Detection Systems can be classified into two categories[3]:

- Host-Based IDS (HIDS) – These are software based products installed on a host computer to analyze and monitor all traffic activities on the system application files and operation system.

- Network-Based IDS (NIDS) – These are found on an entire network to capture and analyze the stream of packets through a network.

A performance comparison of Host-Based IDs and Network-Based IDS is shown in table I.

### B. Intrusion patterns

Intrusion detection systems have been designed such that they can detect and identify threats efficiently. The intrusions come down to two patterns of detection [14]:

- Misuse intrusion – It matches the patterns in the intruder's network traffic. One advantage of misuse detection system is its ability to detect all known threats [14].

- Anomaly intrusion- It is behavioral based intrusion detection system. It observes changes in the normal patterns in a system by building a profile of the system which is being monitored [14], [4].

Many systems combine these two intrusion patterns because of their complementary nature. Due to the problem of false positives, systems based on misuse pattern are commonly used for commercial purposes whilst Anomaly detection is found in research systems [15].

### C. Intrusion Detection Techniques

There are different types of intrusion detection techniques based on the intrusion detection patterns. Below is a description of those commonly used including the ANN, which is being proposed in this paper.

- Statistical analysis - this involves comparing current trends in data to a predetermined set of baseline criteria. It compares the normal behavior of data to the deviations over time. This technique is employed in Anomaly detection system [4], [15].

- Evolutionary algorithm - this technique creates an application path, which provides normal behavioral models. These applications are modelled to detect conditions of normal behavior, error conditions and attempted intrusion by classifying the models based on different conditions [15].

- Protocol verification - this is a technique based on thorough checks of protocol fields and their behavior as compared to established standards. Data considered to have violated the established standard is classified suspicious. This technique has its success in commercial; systems but has a disadvantage of giving false positives for unspecified protocols [15].

- Rule Based - the state transition analysis technique compares data against signatures. Each packet is applied to a finite state machine following transitions until a final state is reached, hence detecting an attack [16].

- ANN - The neurons of the ANN are used to form complex hypotheses; the more neurons, the more complex the hypotheses. Evaluating the hypotheses is done by setting the input nodes in a feed-back process and the event streams are propagated through the network to the output where it is classified as normal or compromised. At this stage the gradient descents is used so as to push the error in the output node back through the network by a back propagation process in order to estimate the error in the hidden nodes. The gradient of the cost – function can thus be calculated [17]–[19]. Neural network system undergoes training in order to learn the pattern created in the system.

## III. ARTIFICIAL NEURAL NETWORK LEARNING PROCEDURE

Artificial Neural Network has two learning procedures.
- Supervised Learning Procedure: In supervised learning, the neural network is provided with labelled training set which learns a mapping from inputs $x$ to outputs $y$, given a labelled set of inputs-output pairs $d = \{(x_i, y_i)\}_{i=1}^{N}$ where $d$ is called the training set and $N$ is the number of training examples. It is assumed that $y_i$ is a categorical variable from some infinite set $y_i \in \{1...C\}$ [20].

The multi–layer perceptron (MLP) is a type of ANN that is trained using supervised learning procedures. The MLP was used in [21] to detect intrusions based on an off-line analysis approach. In a different approach, MLP was used in [22] to detect intrusion on network data comparing its performance with Self-Organizing Maps (SOM).

- Unsupervised Neural Network Learning Procedure: in this learning procedure, the neural network is has an input $d = \{x_i\}_{x=1}^{N}$ that is a set of unlabeled data and you are to find patterns in the data.

SOM is a type of ANN that is trained using unsupervised learning procedure to produce a low dimensional, discretized representation of the input space of training samples called Map.

In this work, MLP architecture with three layers feed-forward Neural Network as show in Fig. 1 was used. The network had a unipolar sigmoid transfer function in each of the hidden and output layers' neurons. A stochastic learning algorithm with a mean square error function was used. Nodes labelled $x_1 \cdots x_6$ have been used in Fig.1 to represent the input units of the neural network where circles labelled "+1" are known as the bias units.

The ANN model has six inputs units (layer $l_1$) three hidden units (layer $l_2$) and one output unit (layer $l_3$) where $l$ denotes layers. The network was trained with feed forward learning algorithm and back ward learning algorithm.

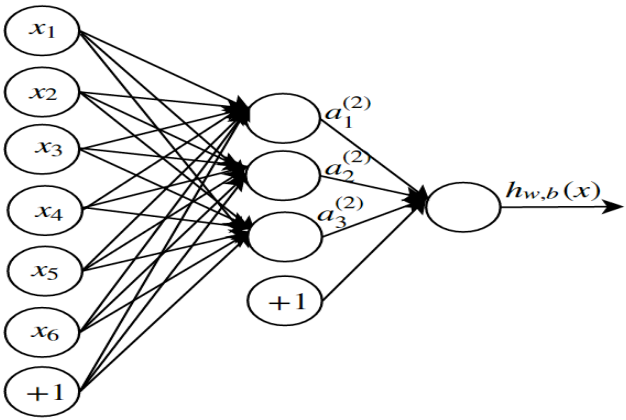

Fig. 1 Three layer Artificial Neural Network

### A. Feed forward Learning Algorithm

Let $s_l$ denote the number of units excluding the bias units. Therefore we make the network parameters $(w,b) = (w^{(1)}, b^{(1)}, w^{(2)}, b^{(2)})$ Where $w_{ij}^{(l)}$ denotes the parameter associated with the connection between unit $j$ in layer $l$ and unit $i$ in layer $l+1$. Also $b_i^{(l)}$ is the bias associated with unit $i$ in layer $l+1$. Thus from the above model, $w^{(1)} \in \Re^{6*3}$ and $w^{(2)} \in \Re^{3*1}$. Let $a_i^1$ denote the output of the unit in layer $l$. For $l=1$ we let $a_i^{(1)} = x_i$ to denote the $i-th$ input. The ANNs model will define a hypothesis $h_{w,b}(x)$ that's outputs a real number.

The model is thus represented mathematically as:

$$a_1^{(2)} = f(w_{11}^{(1)} x_1 + w_{12}^{(1)} x_2 + \cdots + w_{16}^{(1)} x_6 + b_1^{(1)}). \quad (1)$$

$$a_2^{(2)} = f(w_{21}^{(1)} x_1 + w_{22}^{(1)} x_2 + \cdots + w_{26}^{(1)} x_6 + b_2^{(1)}). \quad (2)$$

$$a_3^{(2)} = f(w_{31}^{(1)} x_1 + w_{32}^{(1)} x_2 + \cdots + w_{36}^{(1)} x_6 + b_3^{(1)}). \quad (3)$$

$$h_{w,b}(x) = f(w_{11}^{(2)} a_1^{(2)} + w_{12}^{(2)} a_2^{(2)} + w_{13}^{(2)} a_3^{(2)} + b_1^{(2)}). \quad (4)$$

Eq. (4) which is the weighted sum of the input to unit $i$ in $l$ is the feedforward learning algorithm.

### B. The Backward Learning Algorithm

This learning process goes through four steps as shown below.
- The feedforward-learning algorithm computes the activation for all the layers in the network.

- The out of $l_3$ is set to compute the error term in the output:

$$\delta_i^{(l_3)} = \frac{\partial}{\partial z_i^{(l_3)}} \frac{1}{2} \| y \cdot h_{w,b}(x) \|^2 \quad (5)$$

$$= -(y_i - a_i^{(l_3)}) \cdot f^{'}(z_i^{(l_3)}).$$

Here $a_i^{(3)} = f(z_i^{(l_3)})$ is the sigmoid function.

- We compute the errors for $l=2, l=3$ for each of the nodes $i$ in layer $l$.

$$\delta_i^{(l)} = (\sum_{j=1}^{s_l} w_{ji}^{(l)} \delta_j^{(l+1)}) f^{'}(z_i^{(l_3)}). \quad (6)$$

- Finally the desired partial derivatives are calculated as :

$$\frac{\partial}{\partial b_i^{(l)}} J(w,b;x,y) = a_j^{(l)} \delta_i^{l+1} = \delta_i^{l+1}. \quad (8)$$

Training the ANN involves taking repeated steps of gradient descent to reduce the cost function $J(w,b)$.

## IV. EXPERIMENTAL SCENARIO

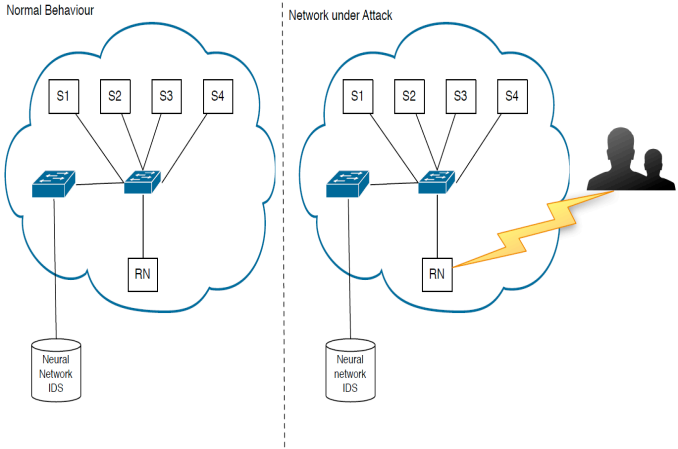

Fig. 2 Experimental architecture

The IoT network is composed of 5 node sensors. Four of the nodes are acting as client, and one is acting as a server relay node for data analytic purposes. The traffic is captured via a network tap avoiding modification of the live traffic.

The server node acknowledges the data sent by the sensor nodes and replies with data based on the received data. This allows the sensor nodes to adapt their behavior and react to occurring events as shown in Fig. 2 (Left).

In the context of this research the attack is from an external intruder. The attack is shown in Fig. 2 (Right). The attackers only target the server node, as it is the one analyzing, logging and responding to the sensors nodes. The DoS attack was performed using a single host, sending over 10 million packets. While the DDoS attacks were performed by up to 3 hosts sending over 10 million packets each at wire speed overflowing the server node. The packets sent during the DoS/DDoS attacks are UDP packet crafted by a custom script in C.

As the server node becomes un-responsive the sensors nodes are not able to adapt their behavior, ultimately leading to a fault on the monitored system.

The detection of the attack is therefore crucial, allowing the response team to avoid disruption of the sensor network and guarantee the stability of the network.

## V. RESULTS AND DISCUSSIONS

This section provides an evaluation of the performance of the ANN intrusion Detection described in section IV. The network was trained with 2313 samples, validated with 496 samples and 496 test samples. Table II shows the number of samples used for classification.

Table II. Number of samples used for classification

| Attack Type | Sample size | Sample Percentage (%) |
|---|---|---|
| DDos/Dos | 2121 | 64.18 |
| Normal | 1184 | 35.82 |

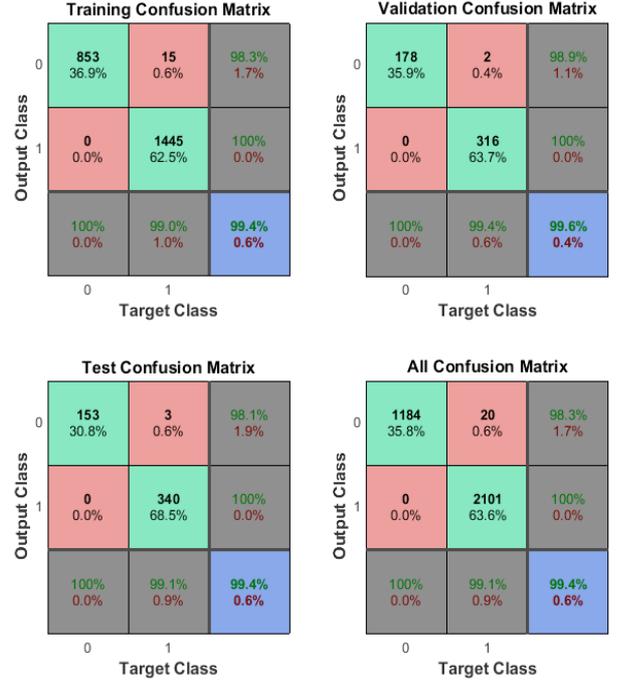

Fig. 3 Neural Network Training confusion matrix

Fig. 3 shows neural network confusion matrix plots for the training set, testing set, validation set and the all confusion matrix (overall performance) at the right bottom corner. The network output correct response values fall under two categories: True positive (TP) and False positive (FP).

TP output gives a measure of attacks rightly classified as shown in green box. FP is a measure of normal events classified rightly as shown in the red box.

The neural network model shows an overall accuracy of 99.4% in classification.

This model demonstrates that the ANN algorithm implemented is able to successfully detect DDoS/DoS attacks against legitimate IoT network traffic. Moreover it helps improving stability of the network by warning the response team at an early stage of the attack, avoiding major network disruptions.

## VI. CONCLUSION AND FUTURE WORK

In this paper, we presented a neural network based approach for intrusion detection on IoT network to identify DDoS/DOS attacks. The detection was based on classifying normal and threat patterns. The ANN model was validated against a simulated IoT network demonstrating over 99% accuracy. It was able to identify successfully different types of attacks and showed good performances in terms of true and false positive rates. For future developments, more attacks shall be introduced to test the reliability of our method against attacks and improve the accuracy of the framework. Furthermore we will investigate other deeper neural networks such as the recurrent and convolutional neural network approach.